\documentclass[]{article}
\usepackage{naacl2001}

\title{A Decision Tree of Bigrams\\is an Accurate Predictor of Word Sense}

\author{Ted Pedersen\\
Department of Computer Science\\
University of Minnesota Duluth
\\Duluth, MN 55812 USA\\
\tt{tpederse@d.umn.edu}} 

\begin{document}
\maketitle      

\begin{abstract}
This paper presents a corpus-based approach to word sense
disambiguation where a decision tree assigns a sense to an ambiguous word
based on the bigrams that occur nearby. This approach is evaluated using
the sense-tagged corpora from the 1998 SENSEVAL word sense disambiguation  
exercise. It is more accurate than the average results reported for 30 of  
36 words, and is more accurate than the best results for 19 of 36 words.
\end{abstract}

\section{Introduction}

Word sense disambiguation is the process of selecting the most appropriate
meaning for a word, based on the context in which it occurs. For our
purposes it is assumed that the set of possible meanings, i.e., the sense 
inventory, has already been determined. For example, suppose {\it bill} 
has the following set of possible meanings: a piece of currency, pending
legislation, or a  bird jaw. When used in the context of {\it The Senate 
bill is under consideration}, a human reader immediately understands that 
{\it bill} is being used in the legislative sense. However, a computer 
program  attempting to perform the same task faces a difficult problem 
since it does not have the benefit of innate common--sense or linguistic 
knowledge. 

Rather than attempting to provide computer programs with real--world 
knowledge comparable to that of humans, natural language processing has 
turned to {\it corpus--based} methods. These approaches use techniques 
from statistics and machine learning to induce models of language 
usage from large samples of text.  These models are trained to perform 
particular tasks, usually via supervised learning. This paper describes an
approach where a {\it decision tree} is learned from some number of 
sentences  where each instance of an  ambiguous word has been manually 
annotated with a sense--tag that denotes the most appropriate sense for 
that context. 

Prior to learning, the sense--tagged corpus must be  converted into a more 
regular form suitable for automatic processing. Each sense--tagged 
occurrence of an ambiguous  word is  converted into a feature vector, 
where each feature represents  some  property of the surrounding text that 
is considered to be relevant to  the disambiguation process.  Given the 
flexibility and complexity of human  language,  there is  potentially an 
infinite set of features that could  be utilized. However, in 
corpus--based approaches features usually consist of information that  can 
be readily identified in the text,  without relying on extensive external 
knowledge sources.  These  typically  include the  part--of--speech of 
surrounding words, the  presence of certain key words within some window 
of context, and various  syntactic properties  of the sentence and the 
ambiguous word. 

The approach in this paper relies upon a  feature set made up of {\it 
bigrams}, two word sequences that occur in a text. The context in which 
an ambiguous word  occurs is  represented by some number of binary 
features that  indicate  whether or  not a  particular bigram has occurred 
within approximately 50 words to the left or right of the word being  
disambiguated. 

We take this approach since surface lexical features 
like bigrams, collocations, and co--occurrences often contribute a great 
deal to disambiguation accuracy. It is not clear how much
disambiguation accuracy is improved through the use of features 
that are identified by more complex pre--processing such as  
part--of--speech tagging, parsing, or anaphora resolution. One of our
objectives is to establish a clear upper bounds on the accuracy of
disambiguation using feature sets that do not impose substantial 
pre--processing requirements. 

This paper continues with a discussion of our methods for identifying 
the bigrams that should be included in the feature set for learning. Then 
the decision tree learning algorithm is described, as are some benchmark 
learning algorithms that are included for purposes of comparison. The 
experimental data is discussed, and then the empirical  results are  
presented. We close with an analysis of our findings and a  discussion of 
related work. 

\section{Building a Feature Set of Bigrams}

We have developed an approach to word sense disambiguation that   
represents text entirely in terms of the occurrence of bigrams, which we 
define to be two consecutive words that occur in a text. The
distributional characteristics of bigrams are fairly consistent across
corpora; a majority of them only occur one time. Given the sparse and
skewed nature of this data, the statistical methods used to select
interesting bigrams must be carefully chosen. We explore two alternatives, 
the power  divergence family of goodness of fit statistics and the Dice 
Coefficient, an information theoretic measure related to pointwise 
Mutual Information. 

Figure \ref{fig:bigram} summarizes the notation for word and bigram counts 
used in this paper by way of a $2 \times 2$ contingency table. The value 
of $n_{11}$ shows how many times the bigram {\it big cat} occurs in the 
corpus. The value  of $n_{12}$ shows how often bigrams occur where  {\it 
big} is the first word  and {\it cat} is not the second. The counts in 
$n_{+1}$ and $n_{1+}$  indicate how  often words {\it big} and {\it cat} 
occur as the first and  second words of any bigram in the corpus. The 
total number of bigrams in the corpus is represented by $n_{++}$. 

\begin{figure}
\begin{center}
\begin{tabular}[c]{@{}r|c|c|@{}l@{}} 
\multicolumn{4}{c}{ } \\
\multicolumn{1}{l}{ } &
\multicolumn{2}{c}{} \\ 
\multicolumn{2}{r}{cat} & 
\multicolumn{1}{c|}{$\neg${cat}} &
\multicolumn{1}{c}{{totals}} \\
\cline{2-4}
big & $n_{11}$=$\hfill$10& $n_{12}$=$\hfill$20& $n_{1+}$=$\hfill$30\\
\cline{2-3}
$\neg${big} & $n_{21}$=$\hfill$40&$n_{22}$=$\hfill$930&$n_{2+}$=$\hfill$970\\
\cline{1-4}
totals& 
\multicolumn{1}{r}{$n_{+1}$=50} & $n_{+2}$=950 & $n_{++}$=1000  \\
\multicolumn{4}{c}{ } \\
\end{tabular}
\caption{Representation of Bigram Counts}
\label{fig:bigram}
\end{center}
\end{figure}

\subsection{The Power Divergence Family}

\cite{CressieR84} introduce the power divergence family of goodness of fit 
statistics. A number of well known statistics belong to this family, 
including the  likelihood ratio statistic $G^2$ and Pearson's $X^2$ 
statistic. 

These measure the divergence of the observed ($n_{ij}$) and expected 
($m_{ij}$) bigram counts, where $m_{ij}$ is estimated based on 
the assumption that the component words in the bigram occur together
strictly by chance:
\begin{eqnarray*}
m_{ij} = \frac{n_{i+} * n_{+j}}{n_{++}}
\end{eqnarray*}

Given this value, $G^2$ and $X^2$ are calculated as:
\begin{eqnarray*}
G^2 = 2 \sum_{i,j} n_{ij} * \log \frac{n_{ij}}{m_{ij}} \ \ \ \ 
\end{eqnarray*}
\begin{eqnarray*}
X^2 = \sum_{i,j} \frac{(n_{ij} - m_{ij})^2}{m_{ij}}
\label{eq:x2}
\end{eqnarray*}     

\cite{Dunning93} argues in favor of $G^2$ over $X^2$, especially when
dealing with very sparse and skewed data distributions.  However,  
\cite{CressieR84} suggest that there are cases where Pearson's statistic 
is more reliable than the likelihood ratio and that one test should not
always be preferred over the other. In light of this,  
\cite{Pedersen96}  presents Fisher's exact test as an alternative since 
it does not rely on the distributional assumptions that underly both  
Pearson's test and the likelihood ratio.  

Unfortunately it is usually not clear which test is most appropriate
for a particular sample of data.  We take the following 
approach, based on the observation that all tests should assign
approximately the same  
measure of statistical significance when the bigram counts in the 
contingency table do not violate any of the distributional assumptions  
that underly the goodness of fit statistics. We perform tests using
$X^2$, $G^2$, and Fisher's exact test for each bigram.  If the 
resulting measures of statistical significance differ, then the  
distribution of the bigram counts is causing at least one of the tests to 
become unreliable. When this occurs we rely upon the value from Fisher's  
exact test since it  makes fewer assumptions about the underlying 
distribution of data. 

For the experiments in this paper, we identified the top 100 ranked 
bigrams that occur more than 5 times in the training corpus associated
with a word. There were no  cases where rankings produced by $G^2$, $X^2$, 
and Fisher's exact test disagreed, which is not altogether surprising 
given that low frequency bigrams were excluded. Since all of these 
statistics produced the same rankings, hereafter we make no distinction
among them and simply  refer to them generically as the power divergence  
statistic. 

\subsection{Dice Coefficient}

The Dice Coefficient is a descriptive statistic that provides a
measure of association among two words in a corpus. It is similar to 
pointwise Mutual Information, a widely used measure that was first 
introduced for identifying lexical relationships in 
\cite{ChurchH90}. Pointwise Mutual Information can be defined as follows: 
\begin{eqnarray*}
MI(w_1,w_2) = log_2 \frac{n_{11} * n_{++}}{n_{+1} * n_{1+}}
\end{eqnarray*}
where $w_1$ and $w_2$ represent the two words that make up the bigram. 

Pointwise Mutual Information quantifies how often two words occur
together in a bigram (the numerator) relative to how often they occur
overall in the corpus (the denominator). However, there is 
a curious limitation to pointwise Mutual Information. A bigram $w_1w_2$  
that occurs $n_{11}$ times in the corpus, and whose component words $w_1$  
and $w_2$ only occur as a part of that bigram, will result in  
increasingly strong  measures of association as the value of $n_{11}$  
decreases. 
Thus, the maximum pointwise Mutual Information in a given corpus
will be assigned to bigrams that occur one time, and whose component words 
never occur outside that bigram. These are usually not the bigrams that
prove most useful for disambiguation, yet they will dominate a ranked
list as determined by pointwise Mutual Information. 

The Dice Coefficient overcomes this limitation, and can be defined as 
follows: 

\begin{eqnarray*}
Dice(w_1,w_2) = \frac{2* n_{11}}{n_{+1} + n_{1+}}
\end{eqnarray*}

When $n_{11} = n_{1+} = n_{+1}$ the value of $Dice(w_1,w_2)$ will be 1 for 
all values $n_{11}$.  When the value of $n_{11}$ is less than either of the 
marginal totals (the more typical case) the rankings produced by the Dice 
Coefficient are similar to those of Mutual Information. The relationship 
between pointwise Mutual Information and the Dice Coefficient is also 
discussed in \cite{SmadjaMH96}. 

We have developed the Bigram Statistics Package to produce ranked lists of
bigrams using a range of tests. This software is written in Perl and 
is freely available from www.d.umn.edu/\~{}tpederse. 

\section{Learning Decision Trees}

Decision trees are among the most widely used machine learning algorithms.
They perform a general to specific search of a feature space, adding
the most informative features to a tree structure as the search proceeds.
The objective is to select a minimal set of features that efficiently 
partitions the feature space into classes of observations and assemble
them into a tree.  In our case, the observations are manually 
sense--tagged  examples of an  ambiguous word in context and the 
partitions correspond to the different possible senses. 

Each feature selected during the search process is represented by 
a node in the learned decision tree. Each node represents a choice
point between a number of different possible values for a feature. 
Learning continues until all the training examples are accounted for
by the decision tree. In general, such a tree will be overly specific
to the training data and not generalize well to new examples. Therefore 
learning is followed by a pruning step where some nodes are eliminated or
reorganized to produce a tree that can generalize to new circumstances. 

Test instances are disambiguated by finding a path through the learned
decision tree from the root to a leaf node that corresponds with the 
observed features. An instance of an ambiguous word is disambiguated by  
passing it through a series of tests, where each test asks if a  
particular bigram occurs in the available window of context. 

We also include three benchmark learning algorithms in this study: the 
majority classifier, the decision stump, and the Naive Bayesian
classifier. 

The {\it majority classifier} assigns the most common sense in the
training data to every instance in the test data. 
A {\it decision stump} is a one node decision tree\cite{Holte93} that is
created by stopping the decision tree learner after the single most
informative feature is added to the tree. 

The {\it Naive  Bayesian classifier} \cite{DudaH73} is based on certain 
blanket  assumptions about the interactions among  features in a 
corpus. There is no search of the feature space performed to build a 
representative model as is the case with decision trees. Instead, all 
features are included in the classifier and assumed to be relevant to the  
task at hand. There is a further assumption that each feature is 
conditionally independent of all other features, given the sense of 
the ambiguous word. It is most often used with a {\it bag of words}  
feature set, where every word in  the training sample is represented by a  
binary feature that indicates  whether or  not it occurs  in the window of  
context surrounding the ambiguous  word. 

We use the Weka \cite{weka} implementations of the C4.5  
decision tree learner (known as J48), the  decision stump, and the Naive  
Bayesian classifier. Weka is written in Java and is freely available from 
www.cs.waikato.ac.nz/\~{}ml. 

\section{Experimental Data}

Our empirical study utilizes the training and test data from the 1998 
SENSEVAL evaluation of word sense disambiguation systems. Ten teams 
participated in the supervised learning portion of this event. 
Additional details about the exercise, including the data and results
referred to in this paper, can be found at the SENSEVAL web site 
(www.itri.bton.ac.uk/events/senseval/) and in \cite{KilgarriffP00}. 

We included all 36 tasks from SENSEVAL for which training and test data 
were provided. Each task requires that the occurrences of a particular 
word in the test data be disambiguated based on a model learned from
the sense--tagged instances in the training data. Some words were used in 
multiple tasks as different parts of speech. For example, there were two 
tasks associated  with {\it bet}, one for its use as a noun and the other 
as a verb. Thus, there are 36 tasks involving the disambiguation of 29 
different words. 

The words and part of speech associated with each task  are shown in Table 
\ref{tab:results} in column 1. Note that the parts of speech are 
encoded as {\it n} for noun, {\it a} for  adjective, {\it v} for verb, and 
{\it p} for words where the part of speech was not provided. The number of 
test and training instances for each task are shown in columns 2 and 
4. Each instance consists of the sentence in which the ambiguous word 
occurs as well as one or two surrounding sentences.  In general 
the total context available  for each ambiguous word is less than 100 
surrounding words. The number of distinct senses in the test data for
each task is shown in column 3. 

\section{Experimental Method}

The following process is repeated for each task. Capitalization and
punctuation are removed from the training and test data. Two feature
sets are selected from the training data based on the top 100 ranked 
bigrams according to the power divergence statistic and the Dice 
Coefficient. The bigram must have occurred 5 or more times to be 
included as a feature. This step filters out a large number of possible
bigrams and allows the decision tree learner to focus on a small number of
candidate bigrams that are likely to be helpful in the disambiguation
process. 

The training and test data are converted to feature vectors where each 
feature represents the occurrence of one of the bigrams that belong in 
the feature set. This representation of the training data is the actual input
to the learning algorithms.  Decision tree and decision stump learning is
performed twice, once using the feature set determined by the power 
divergence statistic and again using the feature set identified by the 
Dice Coefficient. The majority classifier
simply determines the most frequent sense in the training data and
assigns that to all instances in the test data. The Naive Bayesian 
classifier is based on a feature set where every word that occurs 5 or 
more times in the training data is included as a feature.  

All of these learned models are used to disambiguate the test data. The
test data is kept separate until this stage. We employ a fine grained  
scoring method, where a word is  counted as  correctly disambiguated only  
when the assigned sense  tag  exactly matches the true sense tag. No  
partial credit is assigned for near misses.

\section{Experimental Results}

The accuracy attained by each of the learning algorithms is shown in Table 
\ref{tab:results}. 
Column 5 reports the  accuracy of the majority classifier, columns 6 and 7 
show the best and average accuracy reported by the 10
participating SENSEVAL teams. The evaluation at SENSEVAL was
based on precision and recall, so we converted those scores to accuracy by 
taking their product.  However, the best precision and recall may have 
come from different teams,  so the best accuracy shown in column 6 may 
actually be higher than that of  any single participating SENSEVAL 
system. The average accuracy in column 7  is the product of the average 
precision and recall reported for the participating SENSEVAL teams.
Column 8 shows the accuracy of the 
decision tree using the J48  learning algorithm and the
features identified by a power divergence statistic. 
Column 10 shows the accuracy of the decision tree when the Dice 
Coefficient selects the features. Columns 9 and 11 show  the accuracy of 
the decision  stump based on the power
divergence statistic  and the Dice
Coefficient respectively. Finally, column  13 shows the accuracy of the
Naive Bayesian classifier based on a bag of words feature set. 

The most accurate method is the decision tree based on a feature set
determined by the power divergence statistic.  The last line of Table 
\ref{tab:results} shows the win-tie-loss score of the decision tree/power 
divergence method relative to every other method. A win shows it was more 
accurate than the method in the column, a loss means it was less accurate, 
and a tie means it was equally accurate. The decision tree/power
divergence method was more accurate than the best reported SENSEVAL 
results for 19  of the 36 tasks, and more accurate for 30 of the 36 tasks 
when compared to the average reported accuracy. The decision stumps also 
fared well, proving to be more accurate than the best SENSEVAL results for 
14 of the 36 tasks. 

In general the feature sets selected by the power divergence statistic
result in more accurate decision trees than those selected by 
the Dice Coefficient. The power divergence tests prove to be more reliable  
since they account for all possible events surrounding two words  
$w_1$ and $w_2$; when they occur as bigram $w_1w_2$, when $w_1$ or  
$w_2$ occurs in a bigram without the other, and when a bigram consists of 
neither. The Dice Coefficient is based strictly on the event where $w_1$ 
and $w_2$ occur together in a bigram.

There are 6 tasks where the decision tree / power divergence approach is
less accurate than the SENSEVAL average; promise-n, scrap-n, shirt-n, 
amaze-v, bitter-p, and sanction-p. The most dramatic difference 
occurred with amaze-v, where the SENSEVAL average was 92.4\% and the 
decision tree accuracy was 58.6\%. However, this was an unusual task 
where every instance in the  test data belonged to a single sense that 
was a minority sense in the training data.

\begin{table*}
\caption{Experimental Results} 
\label{tab:results}
\begin{center}
\begin{tabular}{crrr|rrrrrrrrr}
\hline
\hline\rule{0pt}{12pt}
(1)  & (2)  & (3)& (4)   & (5)  & (6)  & (7) & (8) & 
(9) & (10) & (11) & (12) \\
 &   & senses      &    &  &  &  & j48 & stump & j48 & stump & naive \\
word-pos & test & in test & train   & maj  & best & avg & pow & pow
& dice & dice & bayes \\[2pt]
\hline
accident-n & 267    &8   & 227     & 75.3 & 87.1 & 79.6 & 85.0
&
77.2 & 83.9 & 77.2 & 83.1 &\\
behaviour-n & 279    &3   & 994     & 94.3 & 92.9 & 90.2 & 95.7 &
95.7 & 95.7 & 95.7 & 93.2 & \\
bet-n & 274    &15  & 106     & 18.2 & 50.7 & 39.6 & 41.8 &
34.5 & 41.8 & 34.5 & 39.3 & \\
excess-n & 186    &8   & 251     & 1.1  & 75.9 & 63.7 & 65.1 &
38.7 & 60.8 & 38.7 & 64.5 & \\
float-n & 75     &12  & 61      & 45.3  & 66.1 & 45.0 & 52.0 &
50.7 & 52.0 & 50.7 & 56.0 & \\
giant-n & 118    &7   & 355     & 49.2  & 67.6 & 56.6 & 68.6 &
59.3 & 66.1 & 59.3 & 70.3 & \\
knee-n & 251    &22  & 435     & 48.2 & 67.4 & 56.0 & 71.3 &
60.2 & 70.5 &60.2 & 64.1 & \\
onion-n & 214    &4   & 26      & 82.7 & 84.8 & 75.7 & 82.7 &
82.7 & 82.7 & 82.7 & 82.2 & \\
promise-n & 113    &8   & 845     & 62.8  & 75.2 & 56.9 & 48.7 &
63.7 & 55.8 & 62.8 & 78.0 & \\
sack-n & 82     &7  & 97       & 50.0  & 77.1 & 59.3 & 80.5 &
58.5 & 80.5 & 58.5 & 74.4 & \\
scrap-n & 156    &14  & 27      & 41.7  & 51.6 & 35.1 & 26.3 &
16.7 & 26.3 & 16.7 & 26.7 & \\
shirt-n & 184    &8   & 533     & 43.5 & 77.4 & 59.8 & 46.7 &
43.5 & 51.1 & 43.5 & 60.9 & \\
amaze-v & 70     &1   & 316     & 0.0  & 100.0& 92.4 & 58.6 &
12.9 & 60.0 & 12.9 & 71.4 & \\ 
bet-v & 117    &9   & 60      & 43.2  & 60.5 & 44.0 & 50.8 &
58.5 & 52.5 & 50.8 & 58.5 & \\
bother-v & 209    &8   & 294     & 75.0 & 59.2 & 50.7 & 69.9 & 
55.0 & 64.6 & 55.0 & 62.2 & \\
bury-v & 201    &14  & 272     & 38.3 & 32.7 & 22.9 & 48.8 &
38.3 & 44.8 & 38.3 & 42.3 & \\
calculate-v & 218    &5   & 249     & 83.9 & 85.0 & 75.5 & 90.8 &
88.5 & 89.9 & 88.5 & 80.7 & \\
consume-v & 186    &6   & 67      & 39.8 & 25.2 & 20.2 & 36.0 &
34.9 & 39.8 & 34.9 & 31.7 & \\
derive-v & 217    &6   & 259     & 47.9 & 44.1 & 36.0 & 82.5 &
52.1 & 82.5 & 52.1 & 72.4 & \\
float-v & 229    &16  & 183     & 33.2 & 30.8 & 22.5 & 30.1 &
22.7 & 30.1 & 22.7 & 56.3 & \\
invade-v & 207    &6   & 64      & 40.1 & 30.9 & 25.5 & 28.0 &
40.1 & 28.0 & 40.1 & 31.0 & \\
promise-v & 224    &6   & 1160    & 85.7 & 82.1 & 74.6 & 85.7 &
84.4 & 81.7 & 81.3 & 85.3 & \\
sack-v & 178    &3   & 185     & 97.8 & 95.6 & 95.6 & 97.8 &
97.8 & 97.8 & 97.8 & 97.2 & \\
scrap-v & 186    &3   & 30      & 85.5 & 80.6 & 68.6 & 85.5 &
85.5 & 85.5 & 85.5 & 82.3 & \\
seize-v & 259    &11  & 291     & 21.2 & 51.0 & 42.1 & 52.9 &
25.1 & 49.4  & 25.1 & 51.7 & \\
brilliant-a & 229    &10  & 442     & 45.9 & 31.7 & 26.5 & 55.9 &
45.9 & 51.1 & 45.9 & 58.1 & \\
floating-a & 47     &5   & 41      & 57.4 & 49.3 & 27.4 & 57.4 &
57.4 & 57.4 & 57.4 & 55.3 & \\
generous-a & 227    &6   & 307     & 28.2 & 37.5 & 30.9 & 44.9 &
32.6 & 46.3 & 32.6 & 48.9 & \\
giant-a & 97     &5   & 302     & 94.8 & 98.0 & 93.5 & 95.9 &
95.9 & 94.8 & 94.8 & 94.8 &\\
modest-a & 270    &9   & 374     & 61.5 & 49.6 & 44.9 & 72.2 &
64.4 & 73.0 & 64.4 & 68.1 & \\
slight-a & 218    &6   & 385     & 91.3 & 92.7 & 81.4 & 91.3 &
91.3 & 91.3 & 91.3 & 91.3 & \\
wooden-a & 196    &4   & 362     & 93.9 & 81.7 & 71.3 & 96.9 &
96.9 & 96.9 & 96.9 & 93.9 & \\
band-p & 302    &29  &1326     & 77.2 & 81.7 & 75.9 & 86.1 &
84.4 & 79.8 & 77.2 & 83.1 & \\
bitter-p & 373    &14  &144      & 27.0 & 44.6 & 39.8 & 36.4 &
31.3 & 36.4 & 31.3 & 32.6 & \\
sanction-p & 431    &7   &96       & 57.5 & 74.8 & 62.4 & 57.5 &
57.5 & 57.1 & 57.5 & 56.8 & \\
shake-p & 356    &36  &963      & 23.6 & 56.7 & 47.1 & 52.2 &
23.6 & 50.0 & 23.6 & 46.6 & \\[2pt]
\hline
\multicolumn{4}{c|} {win-tie-loss (j48-pow vs. X)}  &
\multicolumn{1}{c} {23-7-6} &
\multicolumn{1}{c} {19-0-17} & 
\multicolumn{1}{c} {30-0-6} & 
\multicolumn{1}{c} {}& 
\multicolumn{1}{c} {28-9-3} & 
\multicolumn{1}{c} {14-15-7} & 
\multicolumn{1}{c} {28-9-3} & 
\multicolumn{1}{c} {24-1-11} & \\[2pt]
\hline
\end{tabular}
\end{center}
\end{table*}

\begin{table*}
\caption{Decision Tree and Stump Characteristics} 
\label{tab:stump}
\begin{center}
\begin{tabular}{c|rrr|rrr}
\hline
\hline
\multicolumn{1}{c|}{ } & 
\multicolumn{3}{c}{power divergence} &
\multicolumn{3}{|c}{dice coefficient} \\
(1) & (2) & (3) & (4) & (5) & (6) & (7) \\
word-pos & stump node & leaf/total & features & stump node & leaf/total
&features \\[2pt] 
\hline 
accident-n & by accident & 8/15 & 101 & by accident & 12/23 & 112 \\ 
behaviour-n & best behaviour & 2/3 & 100 & best behaviour & 2/3 & 104 \\ 
bet-n & betting shop & 20/39 & 50 & betting shop & 20/39 & 50 \\ 
excess-n & in excess & 13/25 & 104 & in excess & 11/21 & 102\\ 
float-n & the float & 7/13 & 13 & the float & 7/13 & 13 \\ 
giant-n & the giants & 16/31 & 103 & the giants & 14/27 & 78 \\ 
knee-n & knee injury & 23/45 & 102 & knee injury & 20/39 & 104 \\ 
onion-n & in the & 1/1 & 7 & in the & 1/1 & 7\\ 
promise-n & promise of & 95/189 & 100 & a promising & 49/97 & 107 \\ 
sack-n & the sack & 5/9 & 31 & the sack & 5/9 & 31 \\ 
scrap-n & scrap of & 7/13 & 8 & scrap of & 7/13 & 8 \\ 
shirt-n & shirt and & 38/75 & 101 & shirt and & 55/109 & 101 \\ 
amaze-v & amazed at & 11/21 & 102 & amazed at  &11/21  & 102 \\ 
bet-v  & i bet & 4/7 & 10 & i bet & 4/7 & 10 \\ 
bother-v & be bothered & 19/37 & 101 & be bothered & 20/39 & 106 \\ 
bury-v & buried in & 28/55 & 103 & buried in & 32/63 & 103 \\ 
calculate-v & calculated to  & 5/9 & 103 & calculated to & 5/9 & 103 \\ 
consume-v & on the & 4/7 & 20 & on the & 4/7 & 20 \\ 
derive-v & derived from & 10/19 & 104 & derived from & 10/19 & 104 \\ 
float-v & floated on & 24/47 & 80 & floated on & 24/47 & 80 \\ 
invade-v & to invade & 55/109 & 107 & to invade & 66/127 & 108 \\ 
promise-v & promise to & 3/5 & 100 & promise you  & 5/9 & 106 \\ 
sack-v & return to & 1/1 & 91 & return to & 1/1 & 91 \\ 
scrap-v & of the & 1/1 & 7 & of the & 1/1 & 7 \\ 
seize-v & to seize & 26/51 & 104 & to seize & 57/113 & 104 \\ 
brilliant-a & a brilliant & 26/51 & 101 & a brilliant & 42/83 & 103 \\ 
floating-a & in the & 7/13 & 10 & in the & 7/13 & 10 \\ 
generous-a & a generous & 57/113 & 103 & a generous & 56/111 & 102 \\ 
giant-a & the giant & 2/3 & 102 & a giant & 1/1 & 101 \\ 
modest-a & a modest & 14/27 & 101 & a modest & 10/19 & 105 \\ 
slight-a & the slightest & 2/3 & 105 & the slightest & 2/3 & 105 \\ 
wooden-a & wooden spoon & 2/3 & 104 & wooden spoon & 2/3 & 101 \\ 
band-p & band of & 14/27 & 100 & the band & 21/41& 117\\ 
bitter-p & a bitter & 22/43 & 54 & a bitter & 22/43 & 54 \\ 
sanction-p & south africa & 12/23 & 52 & south africa & 12/23 & 52 \\ 
shake-p & his head & 90/179 & 100 & his head & 81/161 & 105 \\ 
\hline
\end{tabular} 
\end{center}
\end{table*} %

\section{Analysis of Experimental Results}

The characteristics of the decision trees and decision stumps learned for
each word are shown in Table \ref{tab:stump}. Column 1 shows the
word and part of speech. Columns 2, 3, and 4 are based on the
feature set selected by the power divergence statistic while
columns 5, 6, and 7 are based on the Dice Coefficient. Columns 2 and 5 
show the node selected to serve as the decision stump. Columns 3 and 6
show the number of leaf nodes in the learned decision tree relative to the
number of total nodes. Columns 4 and 7 show the number of bigram
features selected to represent the training data. 

This table shows that there is little difference in the decision stump 
nodes selected from feature sets determined by the power divergence 
statistics versus the Dice Coefficient. This is to be expected
since the top ranked bigrams for each measure are consistent, and the
decision stump node is generally chosen from among those. 

However, there are differences between the feature sets selected by the  
power divergence statistics and the Dice Coefficient. These are reflected 
in the different sized trees that are learned based on these feature sets. 
The number of leaf nodes and the total number of nodes for each learned 
tree is shown in columns 3 and 6. 
The number of internal nodes is simply the difference between the
total nodes and the leaf nodes. 
Each leaf node represents the end of
a path through the decision tree that makes a sense distinction. 
Since a bigram feature can only appear once in the 
decision tree, the number of internal nodes represents the number of 
bigram features selected by the decision tree learner. 

One of our original hypotheses was that accurate decision trees of 
bigrams will include a relatively small number of features. This
was motivated by the success of decision stumps in performing
disambiguation based on a single bigram feature. 
In these experiments, there were no decision trees that used all of the  
bigram features identified by the filtering step, and for many words the 
decision tree learner went on to eliminate most of the candidate 
features. This can be seen by comparing the number of internal nodes with 
the number of candidate features as shown in columns 4 or 7.\footnote{For 
most words the 100 top ranked bigrams form the set of candidate features 
presented to the decision tree learner. If 
there are ties in the top 100 rankings then there may be more than 100  
features,  and if the there were fewer than 100 bigrams that occurred more  
than 5 times then all such bigrams are included in the feature set.} 

It is also noteworthy that the bigrams ultimately selected by the decision 
tree learner for inclusion in the tree do not always include those
bigrams ranked most highly by the power divergence statistic or the Dice 
Coefficient. This is to be expected, since the selection of the bigrams 
from raw text is only measuring the association between two words, while 
the decision tree seeks bigrams that partition  instances of the ambiguous 
word into into distinct senses. In particular, the decision tree learner 
makes decisions as to what bigram to include as nodes in the tree using 
the gain ratio, a measure based on the overall Mutual Information
between the bigram and a particular word sense.

Finally, note that the smallest decision trees are functionally equivalent  
to our benchmark methods. A decision tree with 1 leaf node and 
no internal nodes (1/1)  acts as a majority  classifier. A  decision tree 
with  2 leaf nodes and 1 internal node (2/3) has the structure of a 
decision stump.

\section{Discussion}

One of our long-term objectives is to identify a core set of features 
that will be useful for disambiguating a wide class of words using both
supervised and unsupervised methodologies.

We have presented an ensemble approach to word sense disambiguation 
\cite{Pedersen00b} where multiple Naive Bayesian classifiers, each based 
on co--occurrence features from varying sized windows of context, 
is shown to perform well on the widely studied nouns {\it interest} and
{\it line}. While the accuracy of this approach was as good as any
previously published results, the learned models were complex and 
difficult to interpret, in effect acting as very accurate black boxes. 

Our experience has been that variations in learning algorithms 
are far less significant contributors to disambiguation 
accuracy than are variations in the feature set.  In other words, an  
informative feature set will result in accurate disambiguation when used 
with a wide range of learning algorithms, but there is no  
learning algorithm that can perform well given an uninformative or 
misleading set of features.  Therefore, our focus is on developing and 
discovering feature sets that make distinctions among word senses. Our  
learning algorithms must not only produce accurate models, but they 
should also shed new light on the relationships among features and allow  
us to continue refining and understanding our feature sets. 

We believe that decision trees meet these criteria. A wide range of 
implementations are available, and they are known to be robust and 
accurate across a range of domains. Most important, their structure is 
easy to interpret and may provide insights into the relationships that 
exist among features and more general rules of disambiguation. 

\section{Related Work}

Bigrams have been used as features for word sense disambiguation,
particularly in the form of collocations where the ambiguous word is one
component of the bigram (e.g.,  \cite{BruceW94b}, \cite{NgL96}, 
\cite{Yarowsky95}). While some of the bigrams we identify are collocations 
that include the word being disambiguated, there is no requirement that 
this be the case. 

Decision trees have been used in supervised learning approaches to word
sense disambiguation, and have fared well in a number of comparative 
studies (e.g., \cite{Mooney96}, \cite{PedersenB97A}).  In the former they 
were used with the bag of word feature sets and in the latter they were 
used with a mixed feature set that included the part-of-speech of 
neighboring words, three collocations, and the morphology of the ambiguous 
word. We believe that the approach in this paper is the first time that
decision trees based strictly on bigram features have been employed. 

The decision list is a closely related approach 
that has also been applied to 
word sense disambiguation (e.g., \cite{Yarowsky94}, \cite{WilksS98},
\cite{Yarowsky00}). Rather than building and traversing a tree to perform
disambiguation, a list is employed. In the general case 
a decision list may suffer from less fragmentation during learning than 
decision trees; as a practical matter this means that the decision list
is less likely to be over--trained. However, we believe that fragmentation
also reflects on the feature set used for learning.  Ours consists of at 
most approximately 100 binary features. This  results in a relatively 
small feature space that is not as likely to suffer from fragmentation as 
are larger spaces. 

\section{Future Work}

There are a number of immediate extensions to this work. The first is to 
ease the requirement that bigrams be made up of two consecutive words. 
Rather, we will search for bigrams where the component words may be
separated by other words in the text. The second is to eliminate the
filtering step by which candidate bigrams are selected by a power
divergence statistic. Instead, the decision tree learner would consider 
all possible bigrams. Despite increasing the danger of fragmentation, 
this is an interesting issue since the bigrams judged most informative by 
the decision tree learner are not always ranked highly in the filtering
step. In particular, we will determine if the filtering process ever 
eliminates bigrams that could be significant sources of disambiguation 
information. 

In the longer term, we hope to adapt this approach to unsupervised 
learning, where disambiguation is performed without the benefit of sense 
tagged text. We are optimistic that this is viable, since bigram features 
are easy to identify in raw text. 

\section{Conclusion}

This paper shows that the combination of a simple feature set made 
up of bigrams and a standard decision tree learning algorithm 
results in accurate word sense disambiguation. The results of this 
approach are compared with those from the 1998 SENSEVAL word sense    
disambiguation exercise and show that the bigram based decision tree 
approach is more accurate than the best SENSEVAL results for 19 of 36 
words.

\section{Acknowledgments}

The Bigram Statistics Package has been implemented by Satanjeev Banerjee, 
who is supported by a Grant--in--Aid of Research, Artistry and Scholarship 
from the Office of the Vice President for Research and the Dean of the  
Graduate School of the University of Minnesota. We would like to thank  
the SENSEVAL organizers for making the data and results from the 1998  
event freely available. The comments of three anonymous reviewers were  
very helpful in preparing the final version of this paper. A preliminary 
version of this paper appears in \cite{Pedersen01a}. 

%
%


\end{document}